\documentclass[journal]{IEEEtran}
\usepackage{cite}

\ifCLASSINFOpdf
\usepackage[pdftex]{graphicx}
\else
\fi
\usepackage{xcolor}
\usepackage{subcaption}

\usepackage{amsmath}
\usepackage{algorithmic}
\usepackage[ruled, vlined]{algorithm2e}

\usepackage{booktabs}

\begin{document}
%
\title{Deep Reinforcement Learning for the Electric Vehicle Routing Problem with Time Windows}
%
%
%

\author{Bo Lin,
        Bissan Ghaddar,
        Jatin Nathwani,
\thanks{B. Lin is with the Department of Mechanical and Industrial Engineering, University of Toronto, Toronto, ON M5S 3G8 Canada (email: blin@mie.utoronto.ca).}
\thanks{B. Ghaddar is with Ivey Business School, Western University, London, ON N6G 0N1 Canada (email: bghaddar@uwaterloo.ca).}
\thanks{J. Nathwani is with the Department of Management Sciences, University of Waterloo, Waterloo, ON N2L 3G1 Canada (email: jatin.nathwani@uwaterloo.ca).}
}

%
%

\markboth{A draft updated on June 10, 2021}
{Shell \MakeLowercase{\textit{et al.}}: Bare Demo of IEEEtran.cls for IEEE Journals}
%



\maketitle

\begin{abstract}

The past decade has seen a rapid penetration of electric vehicles (EVs) as more and more logistics and transportation companies start to deploy electric vehicles (EVs) for service provision.
In order to model the operations of a commercial EV fleet, we utilize the EV routing problem with time windows (EVRPTW).
In this paper, we propose an end-to-end deep reinforcement learning framework to solve the EVRPTW.
In particular, we develop an attention model incorporating the pointer network and a graph embedding layer to parameterize a stochastic policy for solving the EVRPTW.
The model is then trained using policy gradient with rollout baseline.
Our numerical studies show that the proposed model is able to efficiently solve EVRPTW instances of large sizes that are not solvable with current existing approaches.

\end{abstract}


%
\IEEEpeerreviewmaketitle

\section{Introduction}
%
%
%
%
\IEEEPARstart{E}{lectric} vehicles (EV) have been playing an increasingly important role in urban transportation and logistics systems for their capability of reducing greenhouse gas emission, promoting renewable energy and introducing sustainable transportation system \cite{boulanger2011vehicle, waraich2013plug}. 
To model the operations of logistic companies using EVs for service provision, Schneider et al. proposed the electric vehicle routing problem with time windows (EVRPTW) \cite{schneider2014electric}. 
In the context of EVRPTW, a fleet of capacitated EVs is responsible for serving customers located in a specific region; 
each customer is associated with a demand that must be satisfied during a time window; 
all the EVs are fully charged at the start of the planning horizon and could visit charging stations anytime to fully charge their batteries.
The objective is to find routes for the EVs such that total distance travelled by the fleet is minimized. 

As an NP-hard combinatorial optimization problem (CO), solving the EVRPTW is computationally expensive. 
Schineider et al. \cite{schneider2014electric} developed a variable neighborhood search and tabu search hybrid meta-heuristic (VNS/TS) that is able to effectively solve benchmark instances.
In a later paper \cite{desaulniers2016exact}, Desaulniers et al. proposed exact branch-and-price-and-cut algorithms for four variants of the EVRPTW according to the number of and the type of recharges.
Both algorithms are able to provide high-quality solutions to the EVRPTW benchmark instances introduced in \cite{schneider2014electric}, yet the solution quality and efficiency decrease as the instance size increases. 
In addition, both algorithms have components that rely on the structure and assumptions of EVRPTW, making it difficult to generalize these algorithms to other EVRPTW variants \cite{lin2021electric}.

This research is motivated by an emerging group of literature on utilizing machine learning (ML) to solve CO. For example, ML could be incorporated into the solution processes to improve primal heuristic \cite{khalil2017learning, nazari2018reinforcement}, make branching decisions \cite{khalil2016learning} and generate cuts \cite{tang2020reinforcement} in a branch-and-bound framework.
On improving primal heuristic, previous studies present two typical paradigms: supervised learning and reinforcement learning (RL). 
Supervised learning models, such as the ones presented in \cite{vinyals2015pointer, li2018combinatorial, selsam2018learning} and \cite{prates2019learning}, are trained with solutions provided by existing algorithms. 
Although they could generate near-optimal solutions to the problems they are trained on \cite{vinyals2015pointer} and could generalize to instances from different distributions \cite{selsam2018learning} and of larger sizes than the ones they have seen during training \cite{li2018combinatorial}, supervised approaches are not applicable to most CO problems as obtaining optimal labels for CO instances is computationally expensive \cite{bello2016neural}.

On the other hand, RL models, such as the ones presented in \cite{bello2016neural, zhang1995reinforcement, mittal2019learning, barrett2019exploratory, khalil2017learning, nazari2018reinforcement} and \cite{kool2018attention}, could learn to tackle CO even without optimal labels.
They consider solving problems through taking a sequence of actions similar to Markov decision process (MDP).
Some reward schemes are designed to inform the model about the quality of the actions it made based on which model parameters are adjusted to enhance the solution quality. 
RL has already been successfully applied to various COs such as the travelling salesman problem (TSP), vehicle routing problem (VRP), minimum vertex cover (MVC), maximum cut (MAXCUT) etc.
Despite the difficulty in training deep RL models, it is currently accepted as a very promising research direction to pursue.

The main objective of this research is to develop an RL model to solve EVRPTW. In particular, based on the framework proposed by Nazari et al. \cite{nazari2018reinforcement} for VRP and TSP, we re-define the system state, rewarding schemes as well as the masking policy for EVRPTW.
The original framework in \cite{nazari2018reinforcement} only considers representation of vertex information and does not take into account graph structure as well as global information which is very important in EVRPTW.
To this end, we incorporate the model with a graph embedding component put forward by Dai et al. \cite{dai2016discriminative} to synthesize local and global information of the graph on which the problem is defined.
The model is then trained using the REINFORCE gradient estimator with greedy rollout baseline \cite{kool2018attention}. 

The proposed model is able to efficiently generate good feasible solutions to EVRPTW instances of very large sizes that are unsolvable with any existing methods. 
It, therefore, could be implemented to support large-scale real-time EV fleet operations.
Moreover, the RL model could be incorporated with other solution algorithms as an initialization for meta-heuristics or as a primal heuristic in mixed integer programming (MIP) solvers, which may assist to enhance solution efficiency and quality.
Furthermore, the model has potential to generalize to other variants of EVRPTW through tailoring the rewarding scheme and masking policy.

The remainder of the paper is structured as follows.
We review previous related literature in Section \ref{sec:related_work}, and formally introduce the problem formulation in Section \ref{sec:EVRPTW}.
We then describe the reinforcement learning framework for EVRPTW in Section \ref{sec:prob_formulation} and provide detailed illustration on our methodology in Section \ref{sec:methodology}.
Computational results and analysis about the proposed approach are presented in Section \ref{sec:computational_results}. Finally, we conclude the paper and suggest possible extensions of the proposed method in Section \ref{sec:conclusion}.

\section{Related Work} \label{sec:related_work}

We first review the literature on utilizing ML to solve CO, focusing on developing primal heuristics. Readers are referred to \cite{bengio2020machine} for a more comprehensive modern survey. The application of neural network (NN) to solving CO dates back to the paper by Hopfield and Tank \cite{hopfield1985neural}. 
They define an array representation for TSP solutions. 
In an $n$-city TSP instance, each city $i$ is associated with an $n$-dimensional array $V_i$ whose $j^{th}$ entry $v_{i,j}$ takes a value of $1$ if city $i$ is the $j^{th}$ city along the route and takes $0$ otherwise.
All the city arrays form an $n\times n$ array modeled by $n^2$ neurons. 
Some motion equations were constructed to describe the time evolution of the circuit in the analogy network comprised of the neurons. 
The circuit finally converge to a ``low-energy'' state favoring high quality feasible solutions to the TSP. 
Although the NN proposed in \cite{hopfield1985neural} does not have a learning process, and its performance heavily relies on the choice of model parameters which hinders its scalability and the generalization capability \cite{wilson1988stability}, it stimulated subsequent research efforts on applying NN to solve CO. 

One promising direction is to solve CO by learning a value function to evaluate each possible adjustment in the current solution or action for constructing solutions.
The value function can then be utilized by search algorithms to find good solutions to the target problem.
For example, for a job-scheduling problem of NASA, Zhang et al. \cite{zhang1995reinforcement} parameterize such a value function as an NN that intakes some hand-designed features of the current schedule and outputs the ``value'' of the possible adjustments. 
For CO that is defined on a graph, hand designed features could be replaced by graph embedding networks that synthesize the structure as well as local and global information of the graph.
Khalil et al. \cite{khalil2017learning} use fitted-Q learning to train a graph embedding network (DQN) for action evaluation based on which they greedily decode solutions to target problems including TSP, MVC and MAXCUT.
Other graph embedding examples could be seen in \cite{li2018combinatorial, selsam2018learning, prates2019learning}, though the embedded graph vectors in \cite{selsam2018learning} and \cite{prates2019learning} are fed to NN to predict problem-specific values instead of evaluating actions.

While \cite{zhang1995reinforcement, khalil2017learning} mainly focus on how to construct NN to estimate values of actions, there are some other research concentrating on the decoding process based on the value function.
For the maximum independent set problem, Li et al. \cite{li2018combinatorial} argue that the naive decoding method, i.e. to greedily select the vertex with the highest value, might lead to poor results because there might exist many optimal solutions and each vertex could participate in some of them.
To address the issue, they propose a tree search paradigm supported by the value function enabling the algorithm to explore a diverse set of solutions.
A graph reduction and a local search component were incorporated to enhance solution efficiency and quality.
To further accelerate the search process, Mittal et al. \cite{mittal2019learning} propose a graph convolution network to prune poor vertices and learn the embeddings of good vertices which are then fed to the model of Li et al. \cite{li2018combinatorial} to produce solution set. 
Moreover, Barrett et al. \cite{barrett2019exploratory} proposed the exploratory DQN allowing the algorithm to revise the actions it previously made so as to more comprehensively explore the solution space.

There is another group of research on applying policy-based approaches, which learn policies to directly determine the next action given a system state, to solve CO.
One good example is the pointer network (PN) developed by Vinyals et al. \cite{vinyals2015pointer} for CO, such as TSP and VRP, whose solutions are permutations of the given vertices.
Inspired by the sequence-to-sequence learning \cite{sutskever2014sequence} originally proposed for machine translation, the PN intakes the given vertices and predict a permutation of them.
The PN is trained in a supervised manner with instance-solution pairs generated by an approximate solver.
To generalize the PN to CO for which instance-solution pairs are difficult to obtain, Bello et al. \cite{bello2016neural} used a policy gradient method to train the PN.
The PN is able to efficietly find close-to-optimal solutions to TSP instances with up to $100$ vertices.
Nazari et al. \cite{nazari2018reinforcement} further generalized this method to the VRP whose vertex states change during the decoding process.
Considering that the order of the vertices does not provide any additional information for a VRP solver, they replace the RNN encoder in the PN with element-wise projections of vertex information which accelerates the model implementation.
On the other hand, Kool et al. \cite{kool2018attention} propose a multi-head attention model for the TSP and VRP.
The model is trained using policy gradient with roll-out baseline which is easier to implement in practice than the A3C method utilized by \cite{nazari2018reinforcement}.

Although value-based methods perform well on various CO problems, they do not directly apply to EVRPTW since some vertices (stations and the depot) could appear more than once in a solution.
Given the similarity between the VRP and the EVRPTW, the policy-based framework proposed by Nazari et al. \cite{nazari2018reinforcement} is a better fit to the EVRPTW, yet global information of the system, which is very important for solving EVRPTW, should also be taken into consideration.
Hence, our proposed model is based on the framework of \cite{nazari2018reinforcement} and 
incorporates a graph embedding component proposed by \cite{khalil2017learning} to synthesize the local and global information of the network. 

This research is also related to the stream of literature on applying reinforcement learning in intelligent transportation system.
With a very similar idea, Yu et al. \cite{james2019online} incorporate the Structure2Vec tool \cite{dai2016discriminative} with PN \cite{vinyals2015pointer} to develop a distributed system for solving an online autonomous vehicle routing problem.
Zhao et al. \cite{zhao2020hybrid} extend the work of \cite{nazari2018reinforcement} to VRPTW by revising the masking scheme and adding a local search phase to further improve the solution provided by the attention model.
In \cite{shi2019operating}, Shi et al. propose an RL framework for ride-hailing service provision in a local community, while in \cite{gao2019reinforcement}, Gao et al. employ the idea of RL to build a data-driven cruise control algorithm for the bus transit line connecting New Jersey and Manhattan, New York.
Our proposed approach differs from them in terms of model architecture, training method as well as problem settings.

\section{Problem Definition}
\label{sec:EVRPTW}

\begin{figure}[t]
    \centering
    \includegraphics[width = 3in]{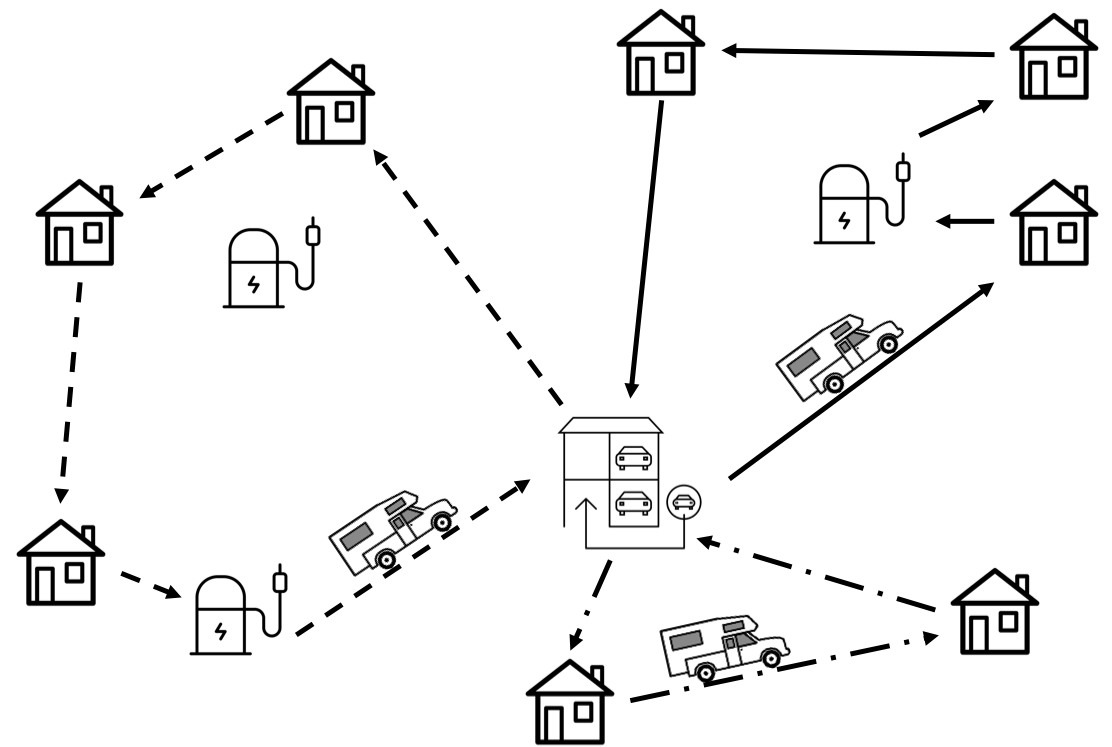}
    \caption{The electric vehicle routing problem with time windows}
    \label{fig:problem_description}
\end{figure}

The EVRPTW proposed by \cite{schneider2014electric} is illustrated in Figure \ref{fig:problem_description}.
We are given a set of customers scattered in a region, each is associated with a demand that need to be satisfied by an EV during a time window.
A fleet of a fixed number of capacitated EVs are initially placed at a depot and are fully charged. 
They could leave the depot to serve the customers and visit stations to recharge their batteries during the planning horizon.
Every time an EV visits a charging station, its battery will be fully charged using linear charging time.
By the end of the planning horizon, they are supposed to return to the depot.
We seek to find routes for the EVs such that all the customer demands are satisfied during their time windows and the total distance travelled by the fleet is minimized.

In order to fit the framework of \cite{nazari2018reinforcement}, we define the EVRPTW on a graph where there are $3$ types of vertices: customer ($V_c$), station ($V_s$) and depot ($V_d$). Each vertex $i$ is associated with an array $X_i^t = \left(x_i, z_i, e_i, l_i, d_i^t \right)$ where $x_i$ and $z_i$ represent the geographical coordinate of vertex $i$, 
$e_i$ and $l_i$ represent the corresponding time window, and $d_i^t$ is the remaining demand at vertex $i$ at decoding step $t$.
The time windows at the stations and the depot are set to $[0, T]$where $T$ is the end of the planning horizon and the demand at these vertices is set to 0.
We superscript $d_i$ and $X_i$ with step $t$ because we solve the problem in a sequential manner, which is introduced in Section \ref{sec:prob_formulation}, and these two elements could change over time. 
All the other elements in $X_i^t$ are static. 
We do not consider the service time at each vertex as \cite{schneider2014electric} because we assume it to be a constant to simplify the problem. 
All the vertex arrays form a set $X^t$ that describes the local information at the vertices at decoding step $t$.
The graph is complete, the weight of each edge is the euclidean distance between the connected vertices.

These nodes share a set of global variables $G^t = \{\tau^t, b^t, ev^t\}$ where $\tau^t$, $b^t$ and $ev^t$ indicate the time, battery level of the active EV and the number of EV(s) available at the start of decoding step $t$ respectively.
The values of $\tau^t$ and $ev^t$ are initially set to $0$ and the size of the fleet respectively. 
The value of $b^t$ is initialized to the EV's battery capacity.
All the global variables could change over time.
We note that, we do not list EV cargo as a global variable here because it is not an input to the model that is introduced in Section \ref{sec:methodology}. But we do keep track on the EV's remaining cargo for the masking scheme implementation.

A solution to the EVRPTW is a sequence of vertices in the graph that could be interpreted as the EVs' routes. 
Routes for different EVs are separated by the depot. 
For instance, suppose vertex $0$ represents the depot, vertex sequence $\{0, 3, 2, 0, 4, 1, 0\}$ corresponds to two routes: one travels along $0\rightarrow 3 \rightarrow 2 \rightarrow 0$, the other one travels along $0\rightarrow 4 \rightarrow 1 \rightarrow 0$, implying that two EVs are used.

\section{Reinforcement Learning for EVRPTW} \label{sec:prob_formulation}

In this section, we describe the problem from a reinforcement learning perspective.
We assume that there is an agent who seeks to generate a solution to the EVRPTW by taking a sequence of actions.
In particular, at each step, the agent intakes the current system state and makes an action based on the given information.
The system state then changes as a consequence.
This procedure is repeated until certain termination conditions are met.
We train the agent with numerous EVRPTW instances and use a reward function to evaluate the solutions generated by the agent and guide the agent to improve accordingly.

In the context of EVRPTW, the system state is the representation of the graph information $X^t$ and $G^t$.
An action is to add (decode) a vertex to the end of the current sequence.
We use $y^t$ to denote the vertex we select at step $t$ and $Y^t$ to denote the vetex sequence we form up to step $t$.
The termination condition is that all the customer demands are satisfied.
We assume the procedure is terminated at step $t_m$.

More specifically, at each decoding step $t$, given $G^t$, $X^t$ and travel history $Y^t$, we estimate the probability of adding each vertex $i$ to the sequence by $P\left(y^{t+1} = i |X^t, G^t, Y^t \right)$, and decode the next vertex to visit, $y^{t+1}$, according to this probability distribution.
Based on $y^{t+1}$, we update the system states using transition functions (\ref{func:tran_time}) - (\ref{func:tran_demand}).

First, system time $\tau^{t+1}$ is updated as follows.

\begin{equation}
    \label{func:tran_time}
    \tau^{t+1} = \left\{
    \begin{aligned}
        & \max(\tau^t, e_{y^t}) + s + w(y^t, y^{t+1})\text{ , if $y^{t}\in V_c$} \\
        & \tau^t + re(b^{t}) + w(y^t, y^{t+1})\text{ , if $y^{t}\in V_s$} \\
        & w(y^t, y^{t+1}) \text{ , if $y^t\in V_d$}
    \end{aligned}
    \right.
\end{equation}
where $w(y^t, y^{t+1})$ is the travelling time from vertex $y^t$ to vertex $y^{t+1}$, $re(b^t)$ is the time required to fully charge the battery from the given level $b^t$, $s$ is a constant representing the service time at each customer vertex.

Next, the battery level of the active EV is updated: 

\begin{equation}
    \label{func:tran_batt}
    b^{t + 1} = \left\{
    \begin{aligned}
        & b^t - f(y^t, y^{t+1})\text{ , if $y^{t} \in V_c$} \\
        & B - f(y^t, y^{t+1})\text{ , otherwise}
    \end{aligned}
    \right.
\end{equation}
where $f(y^t, y^{t+1})$ is the energy consumption of the EV travelling from vertex $y^t$ to vertex $y^{t+1}$, $B$ is the battery capacity.

Finally, the number of EVs available $ev^{t}$, and the remaining demand, $d^t_i$, at each vertex are updated as follows.

\begin{equation}
    \label{func:tran_nev}
    ev^{t + 1} = \left\{
    \begin{aligned}
        & ev^{t} - 1\text{ , if $y^t\in V_d$} \\
        & ev^{t}\text{ , otherwise}
    \end{aligned}
    \right.
\end{equation}

\begin{equation}
    \label{func:tran_demand}
    d_i^{t+1} = \left\{
    \begin{aligned}
        & 0\text{ , $y^t$ = i} \\
        & d^{t}_i\text{ , otherwise}
    \end{aligned}
    \right.
\end{equation}

We define the reward function for a vertex sequence $Y^{t_m} = \{y^0, y^1, \dots, y^{t_m}\}$ as in Equation (\ref{func:reward_traj}).
A high reward value corresponds to a solution of high quality.
Given that the objective of the EVRPTW is to minimize the total distance traveled by the fleet, we set the first term in Equation (\ref{func:reward_traj}) as the negative total distance travelled by the fleet in favor for short-distance solutions.
The other terms are penalties of problem constraint violations. 
If a solution $Y^{t_m}$ requires more than the given EVs, the corresponding $ev^{t_m}$ will be negative which is penalize in the second term.
Moreover, if the depot is located very close to a station, we observe through experiments that the model might achieve low travelling distance by constantly moving between this station and the depot without serving any customers. 
In order to prevent this issue, we introduce the third term to penalize every station visit, which is plausible because we only visit a charging station when necessary under the EVRPTW setting.
In addition, we penalize the negative battery level in the fourth term.
All the other problem constraints are taken into account in the masking scheme introduced in Section \ref{sec:methodology}.

\begin{equation}
\begin{aligned}
    \label{func:reward_traj}
    r(Y^{t_m}) = 
    &-\sum_{t = 1}^{t_m} w(y^{t-1}, y^{t})
    + \beta_{1}\max\{- ev^{t_m}, 0\}\\ 
    &+ \beta_{2} S(Y^{t_m}) + \beta_3 \sum_{t=0}^{t_m} \max\{-b^t, 0\}
\end{aligned}
\end{equation}
where $w(y^{t-1}, y^t)$ is the travelling time on edge ($y^{t-1}$, $y^t$),
$S(Y^{t_m})$ is the number of station visit(s) along trajectory $Y^{t_m}$,
$\beta_{1}$, $\beta_{2}$ and $\beta_3$ are three negative constants. {We note that, according to our experiments, the reward function illustrated in equation \eqref{func:reward_traj} can guide the RL agent to generate solutions subject to the related constraints. However, there is no theoretical guarantee that these constraints will not be violated. If violated, one can consider using method proposed by \cite{zhao2020hybrid} to incorporate a downstream local search heuristic to further improve the solution quality.}

In the next section, we describe the RL methodology in details and explain how it applies to EVRPTW.

\section{Methodology} \label{sec:methodology}

\subsection{The Attention Model}

\begin{figure}[!t]
\centering
\includegraphics[width=3.2in]{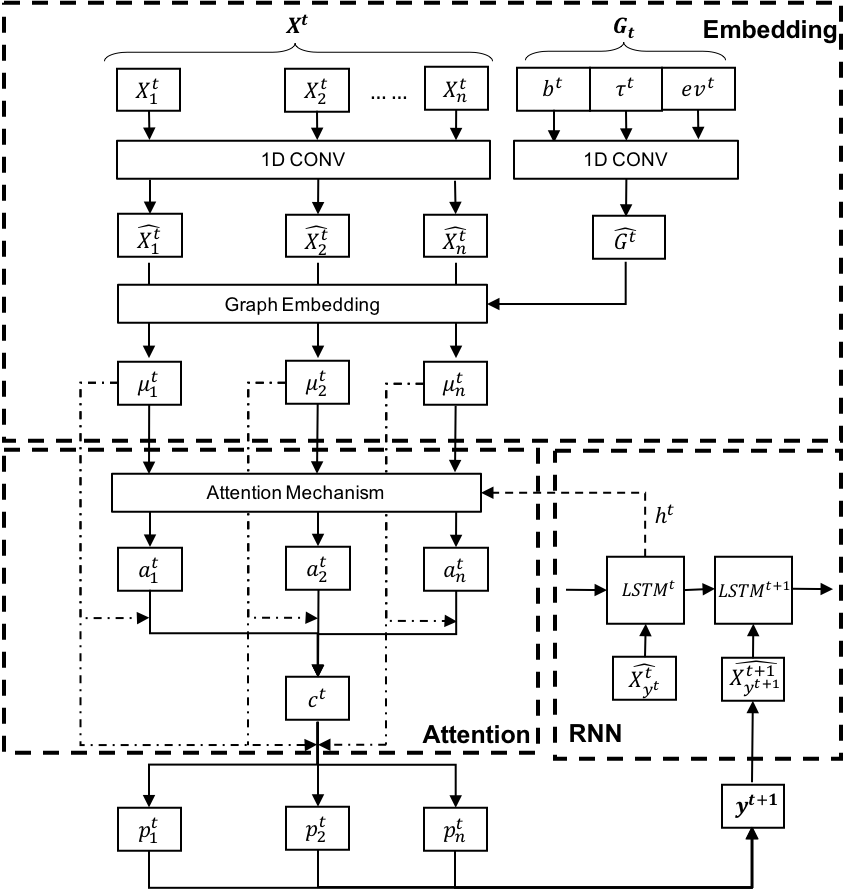}
\caption{The proposed attention model.}
\label{fig:attention_model}
\end{figure}

We propose an attention model to parameterize the "probability estimator", $P(y^{t+1}=i|X^t, G^t, Y^t)$, introduced in the previous section. 
The model consists of $3$ components: an embedding component to represent the system state in a high-dimensional vector form; an attention component to estimate the probability for each vertex; and an LSTM decoder to restore the travel history.
One of the key differences between the proposed model and the model presented in \cite{nazari2018reinforcement} is that we incorporate a graph embedding component to synthesize the local and global information of the graph.
The model structure is illustrated in Figure \ref{fig:attention_model}.

\subsubsection{Graph Embedding}

We first map the model inputs $X^t$ and $G^t$ into a high dimensional vector space.
The embedded model inputs are denoted as $\hat{X^t}$ and $\hat{G^t}$ respectively. 
More specifically, for vertex $i$, its local information array $X_i^t = (x_i, z_i, e_i, l_i, d_i^t)$ is embedded to a $\xi$ dimensional vector $\hat{X_i^t}$ with a $1$-dimensional convolutional layer.
The embedding layer is shared among vertices. 
In addition, we have another $1$-dimensional convolutional layer for global variables $(\tau^t, b^t, ev^t)$, mapping them to a $\xi$-dimensional vector $\hat{G^t}$. 

We then utilize the Structure2Vec tool introduced in \cite{dai2016discriminative} to synthesize the embedded vectors. 
In particular, we initialize a vector $\mu^{(0)}_i = \hat{X_i^t}$ for each vertex $i$, and then update $\mu^{(k)}_i, \forall k = 1, 2, \dots, p$ recursively using Equation (\ref{func:G2V}).
After $p$ rounds of recursion, the network will generate a $\xi$-dimensional vector $\mu^{(p)}_i$ for each vertex $i$ and we set $\mu_i^t$ to $\mu^{(p)}_i$.

\begin{multline}
    \mu^{(k)}_i = relu\{
    \theta_1 \hat{X^t_i} + 
    \theta_2 \hat{G^t}  +
    \theta_3 \sum_{j \in N(i)} \mu^{(k-1)}_j + \\
    \theta_4 \sum_{j\in N(i)} relu\left[\theta_5w(i, j)\right]
    \}
    \label{func:G2V}
\end{multline}
where $N(i)$ is the set of vertices that are connected with vertex $i$ by an edge, we call this set as the neighborhood of vertex $i$, $w(i,j)$ represents the travelling time on edge $(i, j)$, $\theta_1$, $\theta_2$, $\theta_3$, $\theta_4$, and $\theta_5$ are trainable variables. $relu$ is a non-linear activation function, $relu(x) = \max\{0, x\}$.

At each round of recursion, the global information and location information are aggregated by the first two terms of Equation (\ref{func:G2V}), while the information at different vertices and edges propagates among each other via the last two summation terms. 
The final embedded vectors $\mu_i^t$ contains both local and global information, thus could better represent the complicated context of the graph.

\subsubsection{Attention Mechanism}

Based on the embedded vectors $\mu_i^t$, we utilize the context-based attention mechanism proposed by \cite{bahdanau2014neural} to calculate the visiting probability of each vertex $i$. 

We first calculate a context vector $c^t$ specifying the state of the whole graph as a weighted sum of all embedded vectors, as shown in Equation (\ref{func:context_vector}).
The weight of each vertex is defined in Equations (\ref{func:weight_vector}) and (\ref{func:energy_v}).

\begin{equation}
    c^t = \sum_{i = 0}^{|V_c|+|V_s|+1}a^t_i\mu^t_i,
    \label{func:context_vector}
\end{equation}
\begin{equation}
    a_i^t = softmax\left(v^t \right)
    \label{func:weight_vector}
\end{equation}
\begin{equation}
    v_i^t = \theta_v tanh\left(\theta_u \left[\mu_i^t; h^t \right] \right)
    \label{func:energy_v}
\end{equation}
where $v_i^t$ is the $i^{th}$ entry of vector $v^t$, $h^t$ is the hidden memory state of the LSTM decoder, $\theta_v$ and $\theta_u$ are trainable variables, [;] means concatenating the two vectors on the two sides of the symbol ";". $tanh$ is a non-linear activation function, $tanh(x) = \frac{e^x - e^{-x}}{e^x + e^{-x}}$. $softmax$ is the normalized exponential function applied to a vector, $softmax(x)_i = \frac{e^{x_i}}{\sum_k e^{x_k}}$.

Then, we estimate the probability of visiting each vertex $i$ at the next step, $p^t_i$, as in Equations (\ref{func:probability}) and (\ref{func:energy_g}).

\begin{equation}
\label{func:probability}
     p^t_i
     = softmax(g^t)
\end{equation}
\begin{equation}
\label{func:energy_g}
     g_i^t = \theta_g tanh\left( \theta_c[\mu^t_i; c_t] \right).
\end{equation}
where $g_i^t$ is the $i^{th}$ entry of vector $g^t$, $\theta_c$ and $\theta_g$ are trainable variables.

\subsubsection{Masking Scheme}
In order to accelerate the training process and ensure solution feasibility, we design several masking schemes to exclude infeasible routes. 
In particular, suppose that the EV is currently at vertex $i$ at decoding step $t$, if vertex $j, \forall j \neq i$ satisfies one of the following conditions, we assign a very large negative number to the corresponding $v_j^t$ and $g_j^t$ such that the calculated weight $a_j^t$ and probability $p_j^t$ will be very close, if not equal, to $0$:
 
 \begin{itemize}
     \item Vertex $j$ represents a customer, its unsatisfied demand is zero or exceeds the remaining cargo of the EV;
     \item Vertex $j$ represents a customer, the EV's current battery level $b^t$ can not support the EV to complete the trip from vertex $i$ to vertex $j$ and then to the depot;
     \item The earliest arrival time at vertex $j$ violates the time window constraint, i.e. $\tau^t + w(i,j) > l_j$;
     \item If the EV travels to vertex $j$ from vertex $i$ (and recharge at vertex $j$ if it is a station), it will not be able to return to the depot before the end of the planning horizon $T$;
     \item We mask all the vertices except the depot if the EV is currently at the depot and there is no remaining cargo at any customer vertices.
 \end{itemize}

\subsubsection{LSTM Decoder}

Similar to \cite{nazari2018reinforcement}, we use the LSTM to model the decoder network. 
At decoding step $t$, The LSTM intakes the vector representation of the EV's current position $\hat{X_{y^t}^t}$ as well as the memory state from the previous decoding step $h^{t-1}$ and output a hidden state $h^t$ maintaining information about the trajectory up to step $t$, i.e. $Y^t$. 
The memory state $h^t$ is then fed to the attention model as introduced earlier in this section.

\subsection{Decoding Methods}

{
Given the probabilities $p^i_t$, for all vertices $i$ at each decoding step $t$, estimated by the attention model, the agent can decode solutions to an EVRPTW instance. In particular, we consider three decoding strategies as follows.}
{
\begin{itemize}
    \item \textbf{Greedy Decoding:} we greedily select the vertex with the highest probability at each step $t$ as the next vertex to visit, i.e. next vertex $j = arg\,max_i p^t_i$. With this strategy, we generate one solution for each instance.
    \item \textbf{Stochastic Sampling:} we sample the next vertex to visit according to the probability distribution described by $p^t_i$, for all $i$, at each decoding step $t$. We can repeat this procedure to obtain multiple solutions to one instance and report the solution with the shortest distance.
    \item \textbf{Beam Search:} For each instance, we simultaneously maintain multiple solutions with the highest overall probabilities and finally report the best solution among them \cite{neubig2017neural}. Beam search can be regarded as a special greedy strategy, considering the probabilities of solutions instead of transitions.
\end{itemize}}

{
Among these strategies, greedy decoding is the fastest, yet may generate poor solutions due to its myopic nature and the lack of exploration for the solution space. Stochastic sampling and beam search generally achieve a better exploration-exploitation balance, although they might require longer time depending on the number of solutions we generate for each instance.
In this paper, in order to thoroughly explore the solution space, we use the stochastic sampling for modeling training. All the three decoding methods are implemented and compared when testing.}

\subsection{Policy Gradient}

We implement a policy gradient algorithm to train the model.
The basic idea is that, instead of letting the model learn from optimal solutions provided by existing algorithms, we use the reward function defined earlier to evaluate the quality of the solutions generated by the model.
In each training iteration, we use $\theta$ to denote all the trainable variables in Equations \ref{func:G2V}, \ref{func:energy_v} and \ref{func:energy_g}, and $\pi^\theta$ to denote the corresponding stochastic solution policy.
We use $\pi^\theta$ to sample solutions for a batch of $N$ randomly generated instances, and calculate the corresponding rewards. 
Based on the rewards, we estimate the gradient of a loss function with respect to each trainable variable.
We then use the Adam optimizer \cite{kingma2014adam} to update the trainable variables in the model.

When estimating gradients, a good baseline usually reduce training variance and therefore increase speed of learning \cite{kool2018attention}.
Instead of using the A3C methods as in \cite{nazari2018reinforcement} which is difficult to implement in practice, we employ the rollout baseline as proposed by \cite{kool2018attention}.
More specifically, in the first $\Lambda$ training steps, we simply use the exponential moving average of the rewards obtained by the model.
At the $\Lambda^{th}$ step, we set the baseline policy to the policy we have at the end of the $\Lambda^{th}$ step.
After that, we evaluate the baseline policy every $\zeta$ iterations.
We update the baseline policy if and only if the current policy is significantly better than the baseline policy on a seperate test set according to a paired t-test ($\alpha = 5\%$).
We generate a new test set every time the baseline policy is updated.

In particular, we define the key components of the policy gradient method as follows:

\subsubsection{Loss Function}
We aim to minimize the loss function as shown in Equation (\ref{func:loss}).
The loss function represents the negative expected total reward of the trajectory $Y$ sampled using the stochastic policy $\pi^\theta$.

\begin{equation}
\label{func:loss}
    L(\theta) =  - E_{Y \sim \pi^\theta}\left[ r(Y) \right]
\end{equation}

\subsubsection{Gradient Estimation}

We use Equation (\ref{func:gradient_estimator}) to estimate the gradient of the loss function $L(\theta)$ with respect to the trainable variables $\theta$.
The parameter $N$ is the batch size, $X_{[i]}$ is the $i^{th}$ training example in the batch, and $Y_{[i]}$ is the corresponding solution generated using $\pi^\theta$.
Additionally, $BL()$ represents the rollout baseline introduced in \cite{kool2018attention}, and
$P_\theta(Y_{[i]}|X_{[i]})$ indicates the probability of generating solution $Y_{[i]}$ given training example $X_{[i]}$ using stochastic policy $\pi^\theta$. 
We use the probability chain rule put forward by \cite{sutskever2014sequence} to decompose the probability $P_{\theta}(Y_{[i]}|X_{[i]})$ as in Equation (\ref{func:ProbChain}). 
Terms $P_{\theta}(y^{t+1}_{[i]}|X^t_{[i]}, G^t_{[i]}, Y^t_{[i]})$ on the right hand side could be obtained from the model at each decoding step.

\begin{equation}
\label{func:gradient_estimator}
    \nabla_\theta L 
    = \frac{1}{N} \sum_{i=1}^{N}
    \left[ r(Y_{[i]}) - BL(X_{[i]}) \right] \nabla_{\theta} log P_{\theta}(Y_{[i]}|X_{[i]})
\end{equation}
where
\begin{equation}
    \label{func:ProbChain}
    P_{\theta}(Y_{[i]}|X_{[i]}) = \prod_{t=0}^{|Y_{[i]}|- 1}P_{\theta}(y^{t+1}_{[i]}|X^t_{[i]}, G^t_{[i]}, Y^t_{[i]})
\end{equation}

\subsubsection{Instance Generation}

At each training step, we generate $N$ random EVRPTW training instances.
In each instance, the vertices are uniformly distributed among a region $[0, 1]\times [0, 1]$. 
Customer demands are considered discrete, they are randomly selected from $\{0.05, 0.10, 0.15, 0.20\}$ with equal probabilities.
We use a way similar to \cite{solomon1987algorithms} to generate the time window for each customer. 
The center of a time window is uniformly distributed among $[0, 1]$ while the length is normally distributed with mean $0.2$ and standard deviation $0.05$.
The time windows are trimmed, if necessary, to fit the planning horizon $[0, 1]$.
We note that although the feasibility of the instances generated by this method is not guaranteed, according to our experiment, they are actually feasible in most cases.
Since deep learning model in general is robust to random errors in training data, we do not apply any adjustments to those infeasible instances.

We normalize the vehicle specifications in \cite{schneider2014electric} to the interval $[0, 1]$.
Cargo and battery capacities of each EV are set to $1.0$. 
Fully charging an EV from $0$ requires $0.25$ time units. Charging the energy consumed when travelling one unit of distance requires $0.15$ time units.
The planning horizon is $[0, 1]$.
We consider a fleet of $3$ EVs serving $10$ customers in a region with $3$ stations during training. 
We use this small instance size to enhance the instance generation efficiency.
According to our numerical experiments, this does not compromise the model performance.
Test data are generated in the same way as we produce training data, yet the numbers of customers, stations and EVs could vary.

The pseudo code of the training procedure is summarized in Algorithm \ref{alg:training}.

\begin{algorithm}[!t]
    \SetAlgoLined
    initialize the network weights $\theta$,
    and test set $S$\;
    \For{$i = 1, 2,\dots$}{
        generate $N$ random instances $X_{[1]}, X_{[2]}, \dots, X_{[N]}$\;
        \For{$n = 1, 2, \dots, N$}{
        initialize step counter $t_n \gets 0$\;
        \Repeat{termination condition is satisfied}{
        choose $y^{t_n+1}_{[n]}$ according to the probability distribution $P_\theta(y^{t_n+1}_{[n]}|X^{t_n}_{[n]}, G_{[n]}^{t_n}, Y^{t_n}_{[n]})$\;    
        observe new state $X^{t_n+1}_{[n]}, G^{t_n+1}_{[n]}, Y^{t_n+1}_{[n]}$\;
        $t_n \gets t_n + 1$\;
        }
        compute reward $r(Y^{t_n}_{[n]})$\;
        }
        \eIf{$i \leq \Lambda$}{
        $BL(X_{[i]})  \gets avg\left[r(Y_{[1]}^{t_1}), \dots, r(Y_{[N]}^{t_N}) \right]$\; 
        }{
        $BL(X_{[i]}) \gets \pi^{BL}(X_{[i]})$\;
        }
        $\mathrm{d}\theta 
    = \frac{1}{N} \sum_{i=1}^{N}
    \left[ r(Y_{[i]}) - BL(X_{[i]}) \right] \nabla_{\theta} log P_{\theta}(Y_{[i]}|X_{[i]})$\;
    $\theta \gets Adam(\theta, \mathrm{d}\theta)$\;
    \eIf{$i=\Lambda$}{
    initialize baseline $\pi^{BL} \gets \pi^\theta$\; 
    }
    {
        \If{$i \mod \zeta = 0$ and $OneSideTTest\left(\pi^{\theta}(S), \pi^{BL}(S)\right) < \alpha$}{
        $\pi^{BL} \gets \pi^{\theta}$\;
        create new test set S\;
        }
    }
    }
    \caption{REINFORCE with Rollout Baseline}
    \label{alg:training}
\end{algorithm}

\section{Numerical Experiment} \label{sec:computational_results}

\subsection{Experimental Setting}

We perform all the tests using a Macbook Pro (2018) running Mac OS 10.13.6 with 4 CPU processors at $2.3$ GHZ and $16$ GB of RAM. 
The RL model is realized using Tensorflow 2.2.0. The code is implemented in Python.

\begin{table*}[!t]
\caption{Comparison of Average Total Travel Distance of the $5$ Approaches}
\label{tab:total_distance}
\centering

\begin{tabular}{@{}ccccccccccc@{}}
\toprule
           & \multicolumn{2}{c}{CPLEX} & \multicolumn{2}{c}{VNS/TS} & \multicolumn{2}{c}{RL(Stochastic)} & \multicolumn{2}{c}{RL(Greedy)} & \multicolumn{2}{c}{RL(Beam)} \\ \cmidrule(l){2-11} 
Instance   & Distance     & Gap        & Distance      & Gap        & Distance         & Gap             & Distance       & Gap           & Distance      & Gap          \\ \midrule
C5-S2-EV2  & \textbf{2.33}         & 0.00\%     & 2.33          & 0.40\%     & 2.53             & 8.58\%          & 2.67           & 14.59\%       & 2.64          & 13.30\%      \\
C10-S3-EV3 & \textbf{3.64}         & 0.00\%     & 3.64          & 0.85\%     & 4.07             & 11.81\%         & 4.39           & 20.60\%       & 4.38          & 20.33\%      \\
C20-S3-EV3 & -            & -          & \textbf{5.34}          & 0.00\%     & 6.41             & 20.04\%         & 7.27           & 36.14\%       & 7.48          & 40.07\%      \\
C30-S4-EV4 & -            & -          & \textbf{6.87}          & 0.00\%     & 8.46             & 23.14\%         & 9.76           & 42.07\%       & 10.58          & 54.00\%      \\
C40-S5-EV5 & -            & -          & -             & -          & \textbf{11.17}            & 0.00\%          & 12.70          & 13.70\%       & 14.72         & 31.78\%.      \\
C50-S6-EV6 & -            & -          & -             & -          & \textbf{14.32}            & 0.00\%          & 16.46          & 14.94\%        & 18.64         & 30.17\%       \\
C100-S12-EV12 & -            & -          & -             & -          & \textbf{41.53}            & 0.00\%          & 43.01          & 3.56\%        & 58.85         & 41.70\%       \\
\bottomrule
\end{tabular}
\end{table*}

\begin{table*}[!t]
\caption{Comparisons of Average Solution Time of the $5$ Approaches}
\label{tab:sol_time}
\centering
\begin{tabular}{@{}cccccc@{}}
\toprule
Instance   & CPLEX & VNS/TS & RL(stochastic) & RL(Greedy) & RL(Beam) \\ \midrule
C5-S2-EV2  & 0.03  & 1.32   & 2.88           & 0.17       & 0.20     \\
C10-S3-EV3 & 67.65 & 10.37  & 7.63          & 0.35      & 0.40     \\
C20-S3-EV3 & -     & 168.86 & 19.40          & 0.62       & 0.71     \\
C30-S4-EV4 & -     & 536.80 & 43.06          & 1.06       & 1.17     \\
C40-S5-EV5 & -     & -      & 70.26          & 1.69       & 1.86     \\
C50-S6-EV6 & -     & -      & 107.96         & 2.31       & 2.61      \\
C100-S12-EV12 & -     & -      & 401.30         & 7.89       & 8.87      \\
\bottomrule
\end{tabular}
\end{table*}

For the RL model, we adapt most hyper-parameters from the work done by \cite{nazari2018reinforcement}. 
We use two separate $1$-dimensional convolutional layers for the embeddings of local and global information respectively. 
All this information is embedded in a $128$-dimensional vector space. 
We utilize an LSTM network with a state size of $\xi=128$. 
For the Adam optimizer \cite{kingma2014adam}, we set the initial step size to $0.001$, and the batch size to $N = 128$.
To stablize the training, we clip the gradients, $d\theta$, such that their norms are no more than $2.0$. 
With regard to the rollout baseline, we use the moving exponential average baseline in the first $1000$ training steps and evaluate the baseline policy every $100$ training steps after that.
In the reward function, the penalty factors for depot and station visits as well as negative battery level are set to $1.0$, $0.3$ and $100$ respectively.
All the trainable variables are initialized with the Xavier initialization \cite{glorot2010understanding}.
We train the model for $10000$ iterations which takes approximately $90$ hours.

When training the model, we sample the solutions in a stochastic manner to diversify the possible circumstances encountered by the model. 
When testing, we consider all the three decoding methods and compare their performance.
We note that when implementing stochastic decoding for test, we sample $100$ solutions for each instance and report the solution with the shortest total distance.
For beam search, we maintain $3$ solutions simultaneously and report the one with the highest overall probability.

\subsection{Computational Result} 

We compare the performance of three methodologies: CPLEX, the VNS/TS heuristic developed by Schneider et al. \cite{schneider2014electric}, and the proposed reinforcement learning model in Tables \ref{tab:total_distance} and \ref{tab:sol_time}. 

We apply these solution approaches to seven different scenarios whose names indicate the numbers of customers, stations, and available EVs.
For example, ``C5-S2-EV2'' means the scenario of $5$ customers, $2$ charging stations and $2$ EVs.
For each scenario, we solve $100$ instances created in the same way as we produce the training data and report the mean total distance travelled by the EV fleet and the gap with respect to the minimal distance achieved by these algorithms in Table \ref{tab:total_distance}. 
The average solution time in seconds over the $100$ instances in seconds is recorded in Table \ref{tab:sol_time}.
We only report the results for algorithms that can successfully solve an instance within $15$ minutes.

Among the three RL implementation, the stochastic decoding approach always yields solutions with the best quality, though it is more time-consuming than the greedy decoding and beam search.
This finding is consistent with the results presented in \cite{barrett2019exploratory} that learning a policy which directly produces a single, optimal solution is often impractical. 
Instead, exploring the solution space with the stochastic policy usually lead to solutions better than a single ``best-guess''.

On small instances, the proposed approach is able to find feasible solutions efficiently, yet the solution quality is worse than the CPLEX and VNS/TS heuristic. For scenarios ``C5-S2-EV2'' and ``C10-S3-EV3'', the optimality gaps of the best RL implementation (stochastic sampling) are $8.58\%$ and $11.81\%$, respectively, while VNS/TS heuristic and CPLEX can solve the problem to optimality in most cases. 

However, the RL model showcases better scalibility and generalization capability than CPLEX and the VNS/TS heuristic.
When it comes to the scenarios with $20$ or more customers, similar to the results reported in \cite{schneider2014electric}, CPLEX is not able to solve the problem within reasonable time and memory usage. The VNS/TS heuristic outperforms the RL model in terms of solution quality on scenarios "C20-S3-EV3" and "C30-S4-EV4", yet spends $7$-$10$ times the solution time utilized by the RL model. With regards to scenarios with $40$ or more customers, the RL model is the only algorithm that is able to solve the EVRPTW within $15$ minutes. In fact, the RL model only spends on average around $1.8$ minutes to solve instances with $50$ customers.

\begin{figure*}[!th]
    \centering
    \includegraphics[width=6in]{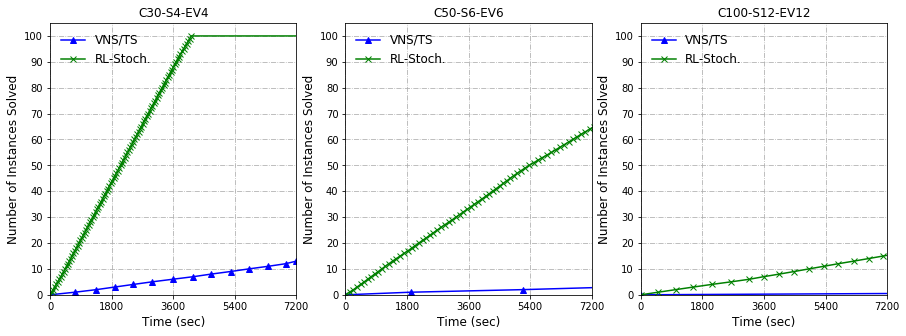}
    \caption{{The number of $100$ given instances solved by the RL model (stochastic sampling) and VNS/TS heuristic in two hours.}}
    \label{fig:num_solved}
\end{figure*}

{
We then lift the $15$-minute solution time limit, and compare the number of instances the VNS/TS heuristic and the RL model (stochastic sampling) can solve within two hours. The results are visualized in Figure \ref{fig:num_solved}. For scenario ``C30-S4-EV4'', the RL model solves all the $100$ given instances in around $40$ minutes, while the VNS/TS heuristics solves only $12$ instances in $2$ hours. The RL agent solves $1300\%$ more instances than the VNS/TS heuristic for scenarios ``C50-S6-EV6''. The VNS/TS heuristic fails to solve any instance in scenario ``C100-S12-EV12'' in $2$ hours, yet the RL model spends on average $4$ minutes to solve an instance. Considering the size of real-world commercial EV fleets, the RL agent is the only approach that can be applicable for large-scale dynamic dispatching. 
}

\subsection{Algorithm Analysis}

\begin{figure*}[!th]
    \centering
    \includegraphics[width=7.5in]{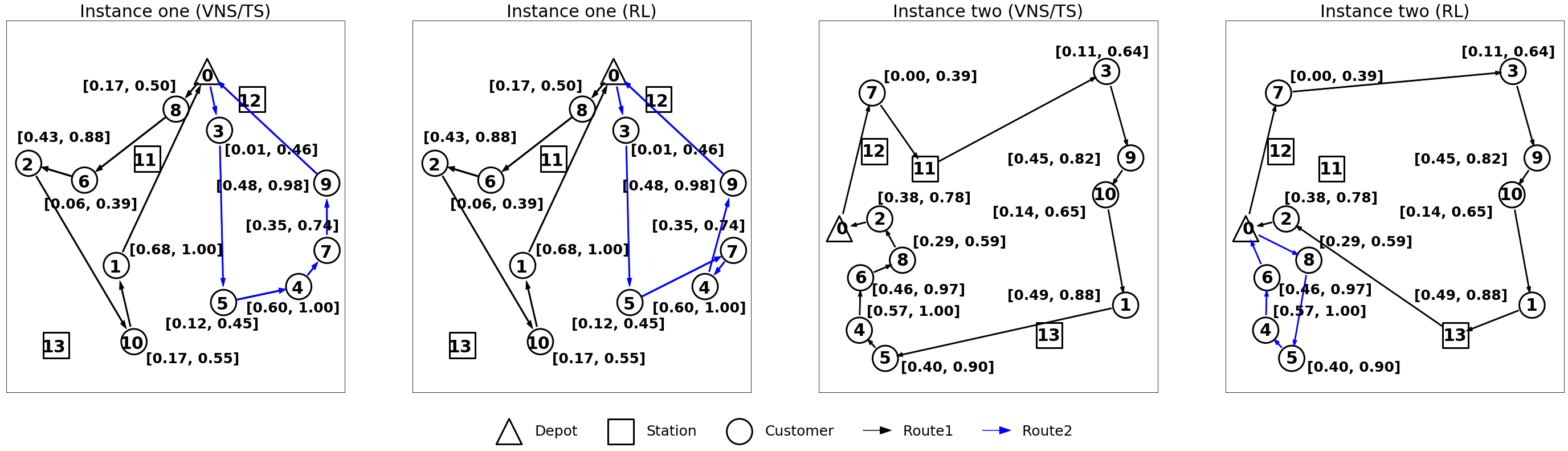}
    \caption{{Sample routes for two instances, each with 10 customers, 3 charging stations and 3 EVs, generated by the VNS/TS heuristic and the stochastic implementation of the RL model, respectively. The brackets beside each customer vertex represent the corresponding time window.}}
    \label{fig:alg_analysis}
\end{figure*}

In this section, we perform detailed analysis on the proposed approach.
Figure \ref{fig:alg_analysis} presents the routes generated by the RL agent with stochastic sampling and the VNS/TS heuristic on two instances, respectively.
Vertices corresponding to customers, charging stations, and the depot are labelled in different colors and shapes. Customer time windows are presented beside the corresponding vertices.

\begin{figure*}[!th]
    \centering
    \includegraphics[width=7.5in]{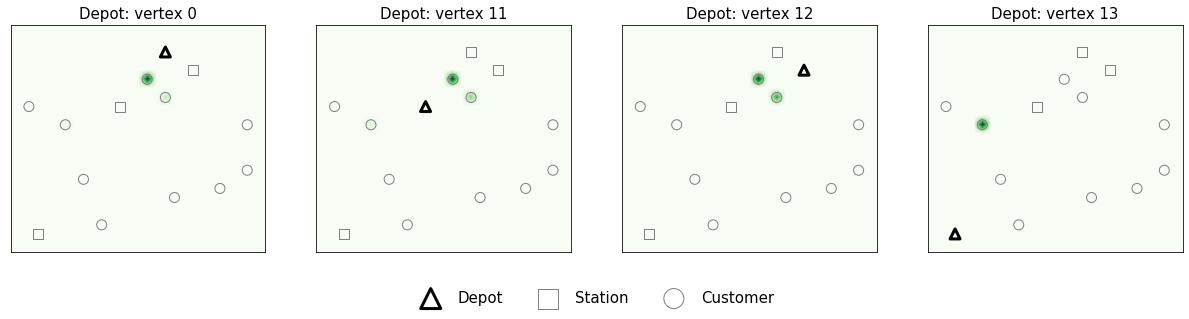}
    \caption{{Visualization of the probability distributions at step $0$ under different depot locations for instance one. The darker a vertex is, the higher the probability of being visited at step $1$.}}
    \label{fig:prob_analysis}
\end{figure*}

One interesting observation from instance one is that the RL agent is able to make the routing decision based on the customers' locations and time windows.
The two EVs both start their route with a vertex (vertices $3$ and $8$) whose time window begins relatively early and close to the depot, and then move to other vertices roughly following the start time of their time windows.
However, there are some exceptions. For example, after serving customer $6$, instead of directly going to customer $10$, it first moves to customer $2$ whose time window starts later than customer $10$ such that the overall travelling distance is reduced. Similar rules apply when considering the order of customers $6$ and $8$.
Nevertheless, the RL agent fails to identify the optimal order of vertices $4$ and $7$ which makes the sole difference compared to the routes generated by VNS/TS heuristic.

{We further perform sensitivity analysis on the depot location for instance one. Figure \ref{fig:prob_analysis} illustrates the probability distributions calculated at step $0$ as we alternate the locations of the depot and charging stations. When the depot is at vertex $0$, the EV is most likely to visit customer $8$ that is closest to the depot, followed by customer $3$ whose time window starts the earliest. As we move the depot to vertices $11$ or $12$, the probability associated with vertex $3$ increases because it becomes closer to the depot. For a similar reason, vertex $6$ is assigned a small probability as we move the depot to vertex $11$. Moreover, when we set vertex $13$ as the depot, vertex $6$ is assigned a very high probability for its early service start time. The RL agent showcases its capability of synthetically considering location and time information to optimize the routing of the EV fleet. The resulting customer sequences, though not necessarily being optimal, are in general of high quality.
}

\begin{figure*}[!th]
    \centering
     \includegraphics[width = 7in]{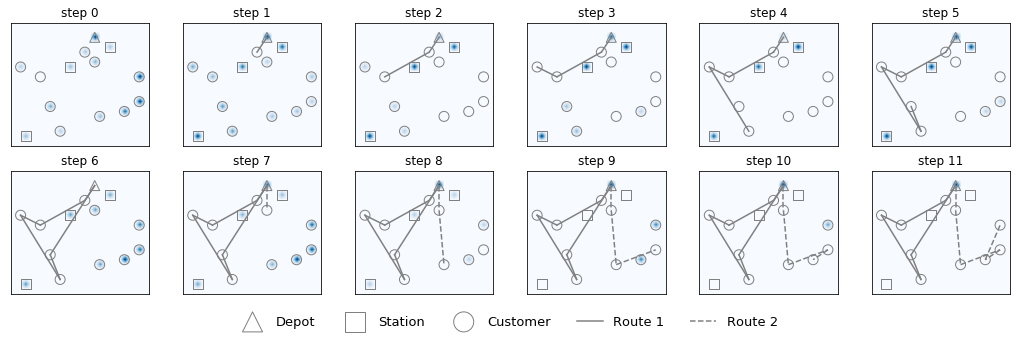}
    \caption{
    {Visualization of the attention mechanism for instance one. The sub-figures present the attention weights the RL agent puts on the vertices. The darker a vertex is, the more attention it receives from the agent.}}
    \label{fig:att_analysis}
\end{figure*}

{
Nevertheless, we also find that the RL agent is sometimes short-sighted, especially for charging decisions. It usually fails to identify charging opportunities early in the trip. Taking instance two in Figure \ref{fig:alg_analysis} as an example, the VNS/TS heuristic utilizes only one EV to serve all the customers while the RL agent needs two EVs.
The reason is that, in the solution generated by the VNS/TS heuristic, the EV charges its battery right after serving customer $7$, taking advantage of the wide time window of customer $3$. As a result, the EV has enough energy for the rest of the trip, and manages to serve all the customers without violating any time windows. Route one generated by the RL agent traverses customers in a very similar order without the detour to station $11$. When the RL agent finally realizes that the EV is short of energy, it sends the EV to station $13$ after serving customer $1$. This detour along with the charging time at station $13$ makes the EV miss the time window of customer $8$. The RL agent thus needs another EV. We also note that the disadvantage of late charging is partially due to the full charging assumption of EVRPTW, i.e. the later the EV charges, the longer charging time it would need.}

{
It is also very interesting to visualize the attention mechanism for the proposed approach. 
Figure \ref{fig:att_analysis} shows the intermediate output $a^t_i$ for all vertices $i$ at each decoding step $t$ for instance one. The darker a vertex is, the greater attention it receives from the RL agent. Throughout the solution processes, the depot along with the charging stations on average receive greater attention compared to the customers. The attention placed on the stations and depot increase as the EVs travel along their routes (from steps $0$ to $6$ and from steps $7$ to $11$, respectively). This trend aligns with our previous observation that the RL agent makes charging decisions mostly based on the battery level of the active EV. The RL agent thus can generate feasible solutions without exhausting an EV's energy, but may fail to find optimal solutions.
}

{
In summary, the proposed RL model is able to capture the structures embedded in the given graph, and combine the location and time information to inform the routing decision makings. The resulting customers' sequences are usually of high quality. With regards to charging, the RL agent makes charging decisions mostly based on EVs' battery levels. It thus ensures that an EV will get charged when it is short of energy, yet may miss some charging opportunities especially at earlier stages. Improvements might be made through developing and training a separate model for charging decisions. Moreover, relaxing the full charging assumption of EVRPTW also showcase an interesting direction for future research.}

\section{Conclusion} \label{sec:conclusion}

In this paper, we developed a reinforcement learning framework for solving the EVRPTW.
Although the solutions generated for small instances by the proposed algorithm are not optimal, we believe it is very promising in practice.
The reasons are three-fold: first, the algorithm showcases great scalability. 
It is able to solve instances of very large sizes which are unsolvable with any existing methods.
Our analysis shows that the proposed model is able to quickly capture important information embedded in the graph, and then effectively provide relatively good feasible solutions to the problem. 
Though not optimal, those good feasible solutions could be utilized to support large-scale real-time EV operations.
Secondly, the proposed model is very efficient in solving the EVRPTW.
In practice, several components of the graph, such as customers' demands and time windows as well as the availability of charging services, could change instantaneously. 
The RL model's ability to efficiently solve the problem allows the EV operators to quickly make adjustments so as to tackle the challenges coming from the stochastic nature of the EVPRTW.
Thirdly, the proposed model can potentially be extended to other variants of the EVRPTW.
Practitioners can extend the proposed method by slightly tailoring the masking schemes as well as the reward function according to their own operational constraints and objectives, which is much easier than adjusting other exact or meta-heuristic algorithms that usually require special assumptions and domain knowledge.

From a theoretical point of view, the proposed solution approach incorporates the graph embedding techniques with the PN architecture, allowing the algorithm to synthesize the local and global information to solve the target problem. We believe its applications are not limited to solving EVRPTW as it could fit with other CO problems that consider both local and global states of the graph on which it is defined.

{
Finally, we highlight several potential extensions of the proposed approach. First, research efforts could be made to design a separate model or a sub-structure in the proposed framework for charging decisions. In doing so, the full charging assumption of EVRPTW might be relaxed to reflect realistic EV operations. Moreover, the solution generated by the RL model could be incorporated into other solution methods, for example, as an initialization method for meta-heuristics, and as a primal heuristic in MIP solvers. In addition, training the RL model with real-world energy consumption and charging data to capture the non-linearity an embedded in the system also present an interesting research direction.}


%



\section*{Acknowledgment}

Bo Lin was supported by the Energy Council of Canada energy policy research fellowship and Bissan Ghaddar was  supported by NSERC Discovery Grant 2017-04185.

\ifCLASSOPTIONcaptionsoff
  \newpage
\fi



\bibliographystyle{IEEEtran.bst}
\bibliography{paper.bbl}
%



%




\end{document}